\documentclass[runningheads]{llncs}
\usepackage{graphicx}
\usepackage{xcolor}         
\usepackage{subfig}
\usepackage{booktabs}       
\usepackage{float}
\usepackage{amsmath}
\usepackage{amsfonts}
\begin{document}
\title{Enhancing Algorithm Performance Understanding through tsMorph: Generating Semi-Synthetic Time Series for Robust Forecasting Evaluation}
\titlerunning{tsMorph: generation of semi-synthetic time series}
%
\author{Moisés Santos\inst{3}\orcidID{0000-0002-1541-8333} \and
André de Carvalho\inst{1} \and
Carlos Soares\inst{2,3}}
\authorrunning{Santos et al.}
%
\institute{Institute of Mathematical and Computer Sciences, University of São Paulo, São Paulo, Brazil \\
\email{andre@icmc.usp.br}
\and
Fraunhofer AICOS Portugal, Porto, Portugal \and
LIACC/Faculdade de Engenharia da Universidade do Porto, Porto, Portugal\\
\email{mrsantos@fe.up.pt, csoares@fe.up.pt}}
\maketitle              
\begin{abstract}
We never produced as much data as today, and tomorrow will probably produce even more data. The increase is due not only to the larger number of data sources but also because the source can continuously produce more recent data. The discovery of temporal patterns in continuously generated data is the main goal in many forecasting tasks, such as the average value of a currency or the average temperature in a city, the next day. In these tasks, it is assumed that the time difference between two consecutive values produced by the same source is constant, and the sequence of values forms a time series. The importance, and the very large number, of time series forecasting tasks make them one of the most popular data analysis applications, which has been dealt with by a large number of different methods. Despite its popularity, there is a dearth of research aimed at comprehending the conditions under which these methods present high or poor forecasting performances. Empirical studies, although common, are challenged by the limited availability of time series datasets, restricting the extraction of reliable insights. To address this limitation, we present tsMorph, a tool for generating semi-synthetic time series through dataset morphing. tsMorph works by creating a sequence of datasets from two original datasets. The characteristics of the generated datasets progressively depart from those of one of the datasets and converge toward the attributes of the other dataset. This method provides a valuable alternative for obtaining substantial datasets. In this paper, we show the benefits of tsMorph by assessing the predictive performance of the Long
Short-Term Memory Network and DeepAR forecasting algorithms. The time series used for the experiments comes from the NN5 Competition. The experimental results provide important insights. Notably, the performances of the two algorithms improve proportionally with the frequency of the time series. These experiments confirm that tsMorph can be an effective tool for better understanding the behavior of forecasting algorithms, delivering a pathway to overcoming the limitations posed by empirical studies and enabling more extensive and reliable experiments. Furthermore, tsMorph can promote Responsible Artificial Intelligence by emphasising characteristics of time series where forecasting algorithms may not perform well, thereby highlighting potential limitations.
\end{abstract}



\section{Introduction}

Forecasting is one of the main tasks of a decision-making process~\cite{Hyndman2018}. Quantitative approaches use historical data, such as time series, to make forecasts~\cite{Montgomery2015}. Time series forecasting is an important tool in several application domains, such as weather, stock markets, and epidemiology. Several methods for time series forecasting have been proposed in the literature with high predictive performance on diverse domains~\cite{Makridakis2018}. However, according to Wang et al. (2022)~\cite{Wang2022}, a limited amount of work tries to understand under which conditions we can expect a forecasting method to obtain good (and bad) results. According to Baeza-Yates et al. (2024)~\cite{baeza2024responsible}, in the context of Responsible AI, being aware of the limitations of methodologies is crucial for mitigating them.

Empirical analysis can be used to understand algorithm behavior. However, sources of realistic datasets for time series analysis benchmarking are limited. Some approaches have been proposed in the literature for generating synthetic time series with realistic characteristics to generate benchmarking and improve the performance of metalearning, such as Autoregressive approaches~\cite{Kang2020}. Autoregressive approaches generate time series with specific characteristics with optimization techniques. Furthermore, generative approaches~\cite{hyland2017real,yoon2019time} based on Generative Adversarial Networks~\cite{goodfellow2020generative} learn the temporal dynamics of realistic time series for the generation of synthetic series. These frameworks focus on the realism of the generated time series. However, generating datasets to improve understanding of the performance of algorithms has two important challenges: (1) the high computational cost of the related approaches; (2) the absence of mechanisms for gradual variation of data characteristics that lead to variation of the behavior of algorithms.

The work of Correia et al. (2019)~\cite{Correia2019} proposed a simple approach for systematic dataset generation called \emph{dataset morphing}. Dataset morphing consists of gradually transforming a source dataset into a target dataset. Gradual changes in the behavior of learning algorithms in the sequence of semi-synthetic datasets aim to obtain a better understanding of these algorithms. This approach was originally proposed for the evaluation of collaborative filtering algorithms. It is a method with an intuitive implementation and, depending on the transformation function, it can have a low computational cost. 
For example, in Correia et al. (2019)~\cite{Correia2019}, a dataset morphing transformation consisted of random rows/columns switching between two collaborative filtering binary datasets.

In this study, we present tsMorph, a novel approach that extends existing work by introducing a dataset morphing technique for generating semi-synthetic time series. Our contributions include:

\begin{itemize}
    \item \textbf{Dataset Morphing for Time Series}: We propose a method, tsMorph, which adapts the dataset morphing approach specifically for time series data.

    \item \textbf{Empirical Algorithm Evaluation}: We demonstrate the application of tsMorph in empirically evaluating forecasting algorithms, providing a valuable contribution to the field. Notably, tsMorph is model-agnostic, allowing seamless integration with various forecasting algorithms.

    \item \textbf{Semi-Synthetic Time Series Generation}: The use of tsMorph enables the creation of semi-synthetic time series with gradual variations, enhancing the versatility of generated datasets for various applications.
\end{itemize}

To validate our methodology, we conducted performance analyses on the Long Short-Term Memory (LSTM) Neural Network and DeepAR algorithms. 
We applied tsMorph to the NN5 competition dataset, providing valuable insights into the effectiveness of these algorithms in time series forecasting. Our findings underscore tsMorph's ability to generate semi-synthetic time series data that capture transitions between datasets. This functionality establishes our approach as a valuable tool for gaining insights into the performance of forecasting algorithms.

\section{Background}

Time series data is characterized by its temporal dependency, where the current value is influenced by past observations. Forecasting models aim to capture and exploit these dependencies to make accurate predictions. In this section, we provide an overview of time series forecasting and performance understanding, highlighting key concepts and techniques in these areas.

\subsection{Time series forecasting}

A time series is a sequence of observations of a variable of interest equally spaced in time~\cite{Montgomery2015}. Then we can denote a time series by $Y = \{ y_{1}, y_{2}, \dots,  y_{T}\}$ and $y_{i} \in \mathbb{R}$ are the observations. The number of observations $T$ is the length of the time series. Such a time series can represent many phenomena in the real world. For example, the demand for beds for patients in a hospital during a period or the daily closing price of a stock in the stock exchange.

One of the main tasks in time series analysis is forecasting. Time series forecasting consists of using the available observations of the variable of interest to extrapolate the time series into the future. Suppose $y$ are the past observations up to period $T$ of a variable of interest such that $y = \{ y_i \}^{T}_{i=0}$. Given $y$, the goal of the time series forecasting task is to obtain a model that estimates the value of the time series at time $T+h$. The estimate is usually represented as  $\hat{y}_{T+h} = y_{T+h} + \epsilon$, for simplicity, $\hat{y}$.

In the forecasting definition, $\hat{y}$ are forecasts from $y$, $h$ denotes the number of forecast observations and is called forecast horizon, and $\epsilon$ is the forecast error. The mapping $F: y \rightarrow \hat{y}$ is a forecasting model. Forecasting models assume a strong relationship between the available observations and the future of a variable of interest~\cite{Montgomery2015}.

To estimate the forecasting accuracy of a model, a set of predictions for the period $T+h$ to $T+h+H$, $\{\hat{y}_{T+h}, \dots, \hat{y}_{T+h+H}\}$, they are compared to the observed values $\{y_{T+h}, \dots, y_{T+h+H}\}$, where $H \in \mathbb{N}^*$.  There are several measures to do this. We use the Mean Absolute Scaled Error (MASE)~\cite{hyndman2006another} in this work. MASE is a scale-free error measure. Because it is scale-free, this measure can be applied to analyze time series from different scales. 

\subsection{Understanding the behavior of ML forecasting algorithms}

In the literature, some approaches are used to understand ML algorithms. This section discusses some that serve as a basis for the proposed approach--starting with dataset morphing and the meta-knowledge analysis for time series forecasting.

Dataset morphing is the process of generating semi-synthetic data from the gradual transformation of a source dataset into a target dataset. Given a source dataset $D^{(s)}$, a target dataset $D^{(t)}$, and a transformation $\tau$, intermediate semi-synthetic datasets $D_{j}$ are obtained by the dataset morphing process. Correia et al. (2019)~\cite{Correia2019} defines a generic dataset morphing process for various tasks in Equation \ref {eq:morph}.

\begin{equation}
\mathcal{D}^{(morph)} : \{D_{j} | D_{0} = D^{(s)}, D_{n} = D^{(t)}, D_{j} = \tau (D_{j-1}) \}, 1 \leq j < n
\label{eq:morph}
\end{equation}

\noindent where $\mathcal{D}^{(morph)}$ is the total set of datasets, and $n$ is the number of transformations. In general terms, the $\tau$ transformation refers to any function applied to the data capable of gradually transforming the source dataset into the target dataset. The dataset morphing process is proposed and used initially to understand the contrasting performance of a pair of algorithms on pairs of datasets. Correia et al. (2019)~\cite{Correia2019} analyzed the performance curves obtained from the $\mathcal{D}^{(morph)}$ datasets. Data characteristics related to performance variation, called meta-features, were also analyzed.

Given a set of datasets $D = \{d_{0}, \dots, d_{n}\}$, a meta-feature m can be defined as a function that maps $m: D \rightarrow R^{ k}$~\cite{alcobacca2020mfe}. Function $m$ returns $k$ values that characterize each dataset in $D$. Several meta-features have been proposed in the literature for different specific tasks. According to Brazdil et al. (2022)~\cite{Brazdil2022}, meta-features must have three main properties: performance discrimination power, computationally not very expensive, and suitable dimensionality to the amount of data available.

Another approach that makes use of meta-features is metalearning. Metalearning is the set of methods that uses knowledge extracted from learning tasks, algorithms, or task performance evaluation to improve predictive performance, make it faster, or understand how algorithms work~\cite{Hutter2019}. It is also occasionally used to analyze the obtained meta-knowledge, despite being typically used for algorithm selection. Here we focus on the use of metalearning.

The work of Armstrong et al. (2001)~\cite{armstrong2001rule} analyzes meta-features in the pioneering work on metalearning for time series forecasting. 
The visualizations presented demonstrate the relationships between meta-features and algorithm selection methods. They serve to assess the benefits derived from automatic selection compared to human expert-based selection.

The approach proposed by Lemke and Gabrys (2010)~\cite{lemke2010meta} uses decision trees to extract meta-knowledge. 
The decision trees were trained using metadata, with the forecasting algorithms serving as the target variable. After inducing a decision tree, rules were extracted on which methodology to follow according to the value of the meta-features.

The work of Talagala et al. (2018)~\cite{talagala2018meta} presents an extensive meta-knowledge analysis. The first analysis consists of a probability matrix of the output of a classification model. The rows represent the time series, the columns are the forecasting models, and the matrix values are the probabilities of selecting a model for a time series. The authors extract interpretations based on the hierarchical clustering of the matrix by columns and the characteristics of the time series in the rows. The second analysis is based on feature importance, measured in individual conditional expectation score (ICE). This analysis aims to measure the effect of changing the value of a single time series feature on the probability output of the algorithm.

Instance Space Analysis (ISA), as elucidated by Smith-Miles and Munoz (2023)~\cite{smith2023instance}, represents a significant shift in how we evaluate algorithms in machine learning. ISA constructs an instance space, mapping all potential test cases onto a two-dimensional plane. This approach uncovers relationships between the structural properties of these cases and algorithm performance. While previous research by Spiliotis et al. (2020)~\cite{spiliotis2020forecasting} applied ISA to assess the suitability of time series data in forecasting competitions, it primarily focused on this aspect and did not explore broader algorithmic analysis or the creation of synthetic time series.

The following section better explores related work to our proposal. In this work, the focus will be on the dataset morphing approach. Dataset morphing is the basis for the development of the tsMorph method.

\section{Synthetic time series}

The need for datasets is a challenge in many applications. A representative diversity of these datasets is also necessary to reach the performance discrimination power and obtain more solid conclusions. In the context of time series analysis, some approaches are available in the literature to generate synthetic data with realistic characteristics. 

The autoregressive approach called GRATIS proposed by \cite{Kang2020} is a method that generates synthetic time series given the desired values of some meta-features. It is an evolutionary approach that searches for the parameters of a data generation method that minimizes the distance between the meta-features of the generated time series and values defined by the user. The data is generated using a Gaussian Mixture Autoregressive (MAR) model.
The applications proposed for GRATIS were generating representative benchmarks and metadata augmentation. 
One important characteristic of this method is that the time series generated may be very different from each other.
This occurs because the similarity that guides the search is calculated in the selected meta-features space and not in the time series space.
Additionally, the search optimizes only some meta-features, which means that the values of other meta-features (i.e., other characteristics of the time series) can be very different.

The paper \cite{yoon2019time} proposed  Time Series Generative Adversarial Networks (TimeGAN), an adaptation of Generative Adversarial Networks (GAN) for time series. GAN is a framework for the generation of realistic data, and in TimeGAN the focus is on the preservation of the temporal dynamics in the generated time series. This is achieved by adding to the classic unsupervised adversarial loss a term representing the similarity to the original data. The work proposes improving the performance of prediction, forecasting, and classification tasks as direct applications.

GRATIS and TimeGAN frameworks are methods for generating synthetic time series that can be used to understand the behavior of algorithms, which is the goal of our work. Both are committed to the realism and diversity of the data generated. However, they have some limitations. First, the time series that is generated may be too different from each other. By analyzing the performance of algorithms on such time series, it is possible to obtain an overall perspective of their behavior (e.g., on what types of time series algorithm A perform better than algorithm B). However, it may also be important to have a more detailed characterization of their behavior (e.g., how does the relative performance of algorithms A and B evolve as the characteristics of the time series gradually change). The time series generated by autoregressive and the generative processes support this kind of analysis. Secondly, the computational costs of TimeGAN training and the GRATIS optimization process are high.

\begin{figure}[htpb]
  \centering
\includegraphics[width=0.7\linewidth]{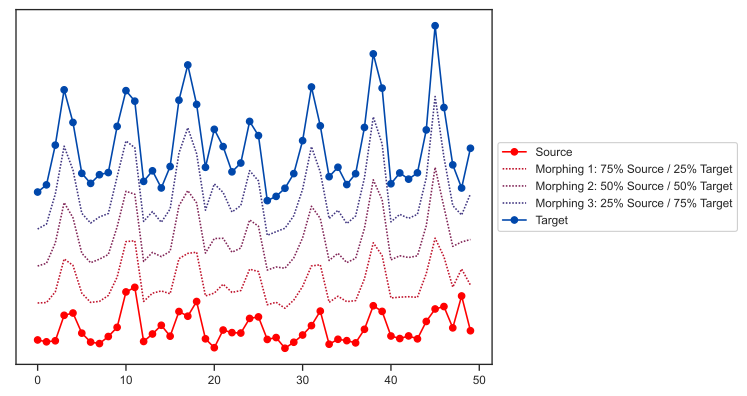}
\caption{Illustrative example of the tsMorph method with n=5.}
\label{fig:tsmorph}
\end{figure}

\section{Dataset morphing for time series}

tsMorph generates semi-synthetic time series for understanding forecasting algorithms. Let a source time series $Y^{(source)}$ and a target time series $Y^{(target)}$ of the same length. The proposed transformation function $\tau$ for the gradual transition between $Y^{(source)}$ and $Y^{(target)}$ is:

\begin{equation}
\tau(i) = \alpha_{i} \cdot Y^{(target)} + ( 1- \alpha_{i} ) \cdot Y^{(source)}, 0 \leq i \leq n
\label{eq:phi_trans}
\end{equation}

\noindent where $\alpha$ is a contribution coefficient equal to $\frac{i}{n-1}$, $n$ is the number of time series after morphing process and $i$ is the index of transformations between $Y^{(source)}$ and $Y^{(target)}$. 
The function $\tau$ is a linear transformation whose contribution coefficient $\alpha$ spaces the generated semi-synthetic time series equally in the range of values between $Y^{(source)}$ and $Y^{(target)}$ during the morphing process.
The computational cost of the $\tau$ transformation is $O(T \times n)$. For time series with length $T >> n$, the computational cost of the transformation $\tau$ is $O(T)$. The dataset morphing for time series developed in the tsMorph method is defined by:

\begin{equation}
Y^{(morph)} : \{Y_{i} | Y_{i} = \tau (i) \}, 1 < i < n
\label{eq:tsmorph}
\end{equation}

\noindent where $Y^{(morph)}$ is the set of datasets and $n$ is the number of time series in the set $Y^{(morph)}$. Therefore, $Y_{i}$ are the semi-synthetic time series gradually generated by the tsMorph method including source and target time series. Figure \ref{fig:tsmorph} illustrates the application of the tsMorph method on source and target time series with $n = 5$.

Figure \ref{fig:tsmorph} illustrates how tsMorph generates a semi-synthetic time series. The value of $\alpha$ for each series can be found in the legend (percentage associated with the source). We can observe that the series are gradually less similar to the source and more similar to the target.

Given $Y^{(source)}$ and $Y^{(target)}$ and one forecasting algorithm, It is possible to understand the behavior of algorithms from the perspective of performance variation and meta-features in the set $Y^{(morph)}$ using tsMorph. We can also control the metadata augmentation process with the generation of semi-synthetic data based on realistic $Y^{(source)}$ and $Y^{(target)}$ time series. 


The advantage of the tsMorph method is the simplicity of the method, which can be easily implemented in any programming language from the formulation. It is a computationally cheap method, as explained earlier. Moreover, it promotes the gradual transformation between the source and target time series. This last characteristic is of great interest in understanding the behavior of forecasting algorithms. 

Traditional evaluation methods in machine learning often focus on performance metrics that summarize algorithm performance across specific datasets, lacking detailed insights into algorithm behavior~\cite{mcgeoch2012guide}. We propose the tsMorph method as a solution for empirical algorithm evaluation in response to this limitation. Unlike conventional approaches, tsMorph aims to understand algorithm performance within a meta-feature space comprehensively. It is crucial to emphasize that, during the morphing process, the application domain of the time series becomes irrelevant, as the primary objective is to identify potential limitations in terms of meta-features inherent to forecasting algorithms when applied to semi-synthetic time series generated from real-world data.

Our method complements other synthetic time series generation methods, such as GRATIS~\cite{Kang2020} and TimeGAN~\cite{yoon2019time}, rather than replacing them. The main focus will be on investigating whether it is possible to extract interesting meta-knowledge from the time series of the set $Y^{(morph)}$ generated with tsMorph to investigate algorithm bias.

\section{Empirical validation}

\subsection{Experimental setup}

We carried out experiments to illustrate the potential of tsMorph to support a better understanding of the performance of forecasting algorithms. The repository for reproducing the results of this study is publicly available~\footnote{Repository URL: \url{https://github.com/moisesrsantos/tsmorph$\_$aequitas}}.

We use the 111 time series from the NN5~\cite{nn52008} Competition. The dataset has years of historical data from cash machines located in different UK regions.
 The goal of the competition is to forecast 56 days. The main characteristic of choosing this set of this dataset is that it is a benchmark known in the literature with a time series of the same size. The data have missing observations that were linearly interpolated since this work objective is not to evaluate this effect.

The forecasting algorithms analyzed in this study include the LSTM and the DeepAR algorithm. LSTM, proposed by Hochreiter and Schmidhuber (1997)~\cite{hochreiter1997long}, is a well-established 
ML algorithm widely employed for time series forecasting tasks. In addition to LSTM, we also investigate the DeepAR algorithm, a probabilistic forecasting model with autoregressive recurrent networks, which represents an advancement beyond LSTM. DeepAR was introduced by Salinas et al. (2020)~\cite{salinas2020deepar} and offers promising capabilities for time series forecasting tasks. These algorithms were chosen for analysis due to their widespread use and the need for a deeper understanding of their behavior. Both algorithms are included in the Python package \texttt{neuralforecast}~\cite{olivares2022library_neuralforecast}.

To illustrate the usefulness of tsMorph, we selected source and target times series from NN5 as follows. We selected ten source time series and one target time series. The ten source time series are the ones where the algorithm obtained the best predictive performance, and the target time series is where the algorithm obtained the worst predictive performance according to MASE. We paired each of the sources with a target and applied tsMorph to generate the corresponding semi-synthetic data. The goal is to understand the changes in the properties of the time series as the predictive performance of the algorithm degrades.

The meta-features used in this work are from the Python package \texttt{catch22}~\cite{lubba2019catch22}. This is a package designed to identify 22 canonical meta-features from time series data. These 22 time series characteristics are carefully chosen from a comprehensive pool of 7000 features within the \texttt{hctsa}~\cite{fulcher2013highly}. Termed as canonical features, they serve as a condensed representation of the larger feature set, with empirical emphasis on optimizing predictive accuracy, computational efficiency, and interpretability. It should be noted that the meta-features were extracted only from the training data, and the performance is extracted only from the test data, while the tsMorph is applied to the complete time series. The meta-feature names used in this work were the short names~\footnote{Short names and descriptions: https://time-series-features.gitbook.io/catch22/feature-descriptions/feature-overview-table}.

\subsection{Understanding the performance of forecasting algorithms}

We delve into the application of tsMorph, a semi-synthetic data generation technique aimed at enhancing our understanding of algorithm performance, particularly for LSTM and DeepAR forecasting algorithms. By leveraging tsMorph, we aim to gain insights into the relationship between algorithm performance and meta-features through augmented time series data.

To assess the impact of tsMorph on enhancing our understanding of algorithm performance, we analyze the correlation between the MASE and meta-features. Specifically, we compute the mean and standard deviation of the correlation across the three meta-features with the lowest standard deviation of correlation for each algorithm.

Table~\ref{tab:cor} presents the results, featuring two sub-tables corresponding to LSTM (a) and DeepAR (b), respectively. Within each sub-table, the mean and standard deviation of the MASE correlation for the selected meta-features are displayed. 

\begin{table}[htpb]
\caption{Pearson correlation between performance and meta-features for the algorithms LSTM (avg MASE = 0.66) and DeepAR (avg MASE = 0.68)}~\label{tab:cor}
\begin{tabular}{c c}
  \begin{tabular}{lrr}
    \toprule
    & \multicolumn{2}{c}{Pearson correlation} \\
    \cmidrule{2-3}
    meta-feature & mean & std \\
    \midrule
    centroid frequency & 0.96    & 0.01    \\
    forecast error & -0.96    & 0.01    \\
    low frequency power & -0.85    & 0.03    \\
    \bottomrule
  \end{tabular}%
&
  \begin{tabular}{lrr}
    \toprule
    & \multicolumn{2}{c}{Pearson correlation} \\
    \cmidrule{2-3}
    meta-feature & mean & std \\
    \midrule
    centroid frequency & 0.99    & 0.00    \\
    forecast error & -0.99    & 0.00    \\
    whiten timescale & -0.91    & 0.01    \\
    \bottomrule
  \end{tabular} \\
  (a)LSTM & (b)DeepAR
\end{tabular}
\end{table}

Tables \ref{tab:cor} presents the Pearson correlation analysis between the MASE and selected meta-features for LSTM and DeepAR algorithms, respectively. Before we interpret the tables, we explain each of these meta-features. A brief description of each of them follows:

\begin{itemize}
\item \textbf{Forecast Error}: this feature provides a measure of the discrepancy arising from employing the mean of the preceding 3 values in the time series to forecast the subsequent value. Time series that are straightforward to predict, signifying instances where the mean of the 3 preceding time steps serves as an accurate prediction for the current value, will yield low values for this feature.
\item \textbf{Centroid Frequency}: this feature calculates the frequency at which the amount of power in frequencies low and higher is the same. It assigns low values to time series with a concentration power in the low frequencies and the opposite for high values.
\item \textbf{Low Frequency Power}: this feature calculates the relative power within the lowest $20\%$ of frequencies. It assigns high values to time series with significant power in low frequencies and low values to time series that predominantly exhibit power in higher frequencies.
\item \textbf{Whiten Timescale}: this feature involves computing the ratio of the first zero-crossing of the autocorrelation function for the residuals to that of the original time series. This ratio provides insight into the relative predictability and stability of the data.
\end{itemize}

For LSTM, the meta-feature "centroid frequency" exhibits a strong positive correlation, while "forecast error" and "low frequency power" show strong negative correlations. Conversely, for DeepAR, similar strong correlations are observed for "centroid frequency" and "forecast error", with an additional negative correlation observed for "whiten timescale". Comparing the two algorithms, DeepAR demonstrates slightly stronger correlations overall, particularly with the "forecast error" meta-feature. This suggests that DeepAR may have a more precise forecasting capability compared to LSTM. The consistent patterns observed in both tables underscore the significance of these meta-features to understanding algorithm performance, providing valuable insights into the behavior of LSTM and DeepAR algorithms in time series forecasting tasks.

\begin{figure}[htpb]
    \centering
    \subfloat[Centroid Frequency.]{\includegraphics[width=0.5\linewidth]{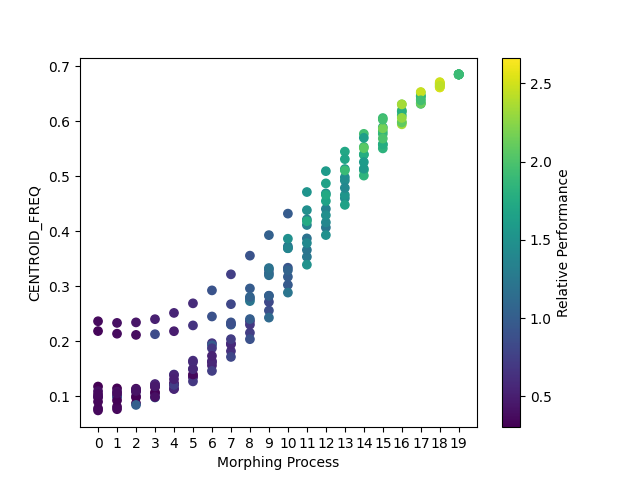} \label{subfig:mf1_lstm}}
    
    \subfloat[Forecast Error.]{\includegraphics[width=0.5\linewidth]{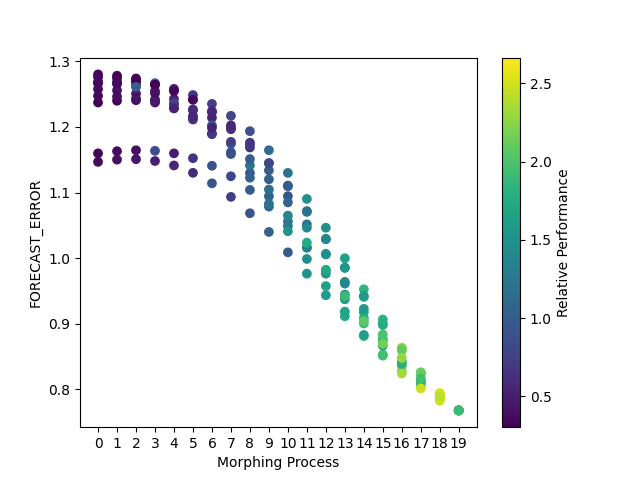} \label{subfig:mf2_lstm}}
    \subfloat[Low Frequency Power.]{\includegraphics[width=0.5\linewidth]{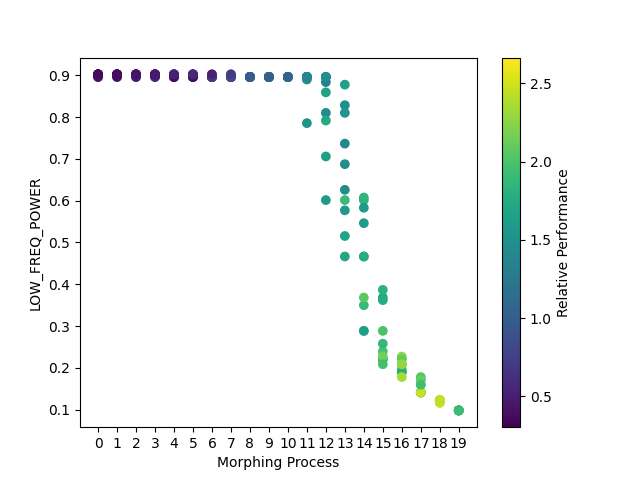} \label{subfig:mf3_lstm}}
 \caption{tsMorph performance understanding plot for LSTM.}
    \label{fig:lstm}
 \end{figure}

\begin{figure}[htpb]
    \centering
    \subfloat[Centroid Frequency.]{\includegraphics[width=0.5\linewidth]{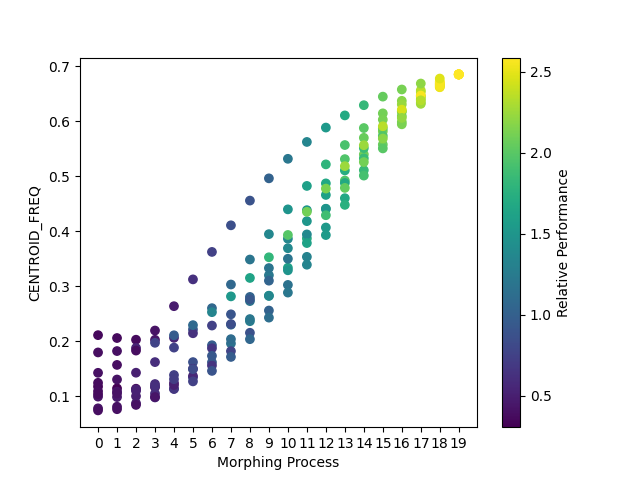} \label{subfig:mf1_deepar}}
    
    \subfloat[Forecast Error.]{\includegraphics[width=0.5\linewidth]{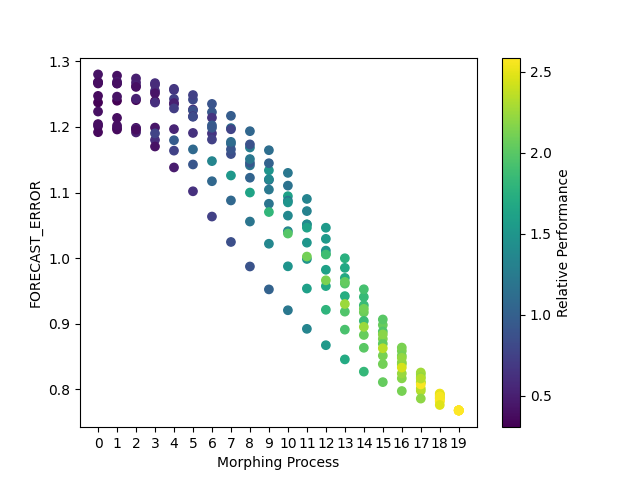} \label{subfig:mf2_deepar}}
    \subfloat[Whiten Timescale.]{\includegraphics[width=0.5\linewidth]{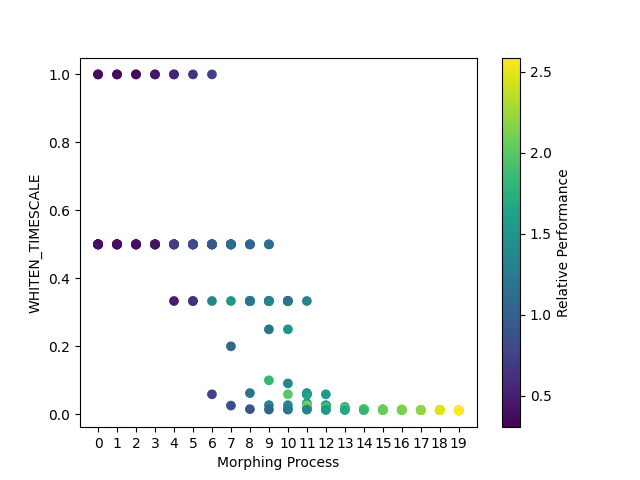} \label{subfig:mf3_deepar}}
 \caption{tsMorph performance understanding plot for DeepAR.}
    \label{fig:deepar}
 \end{figure}

To visually represent the effects of the algorithm performance and meta-feature relationship, we present two sets of figures, each containing three subfigures corresponding to a specific algorithm. Figure~\ref{fig:lstm} depicts the tsMorph performance understanding plot to LSTM, with each subfigure showing the dispersion between meta-feature values on the y-axis, the morphing process step on the x-axis, and a color bar on the right side representing the relative performance in terms of Mean Absolute Scaled Error (MASE). Similarly, Figure \ref{fig:deepar} illustrates the tsMorph performance understanding plot to DeepAR, with each subfigure providing insights into the dispersion of meta-feature values, morphing process steps, and relative performance in MASE.

The analysis of the results reveals that tsMorph was successful in generating semi-synthetic time series with diversity in both the meta-feature space and performance. This achievement is a significant outcome of the study, as it demonstrates the versatility and effectiveness of tsMorph in augmenting time series data for machine learning tasks.
Across both LSTM and DeepAR, a strong negative correlation is observed with the "Forecast Error" meta-feature. This suggests that both algorithms perform better when discrepancies between predicted and actual values are minimized, highlighting their sensitivity to prediction accuracy.
Furthermore, both algorithms exhibit a positive correlation with the "Centroid Frequency" meta-feature, indicating their effectiveness in capturing long-term trends present in high-frequency components of the data. However, while LSTM shows a negative correlation with "Low Frequency Power," suggesting potential challenges with time series exhibiting significant power in low frequencies, DeepAR demonstrates a negative correlation with the "Whiten Timescale" meta-feature. This implies that DeepAR may excel in producing stable and predictable forecasts when the data exhibits lower autocorrelation in the residuals.

\section{Conclusion}

In contrast to the extensive work on developing and testing forecasting algorithms, there is limited work on understanding their behavior. In this work, we propose a simple method of generating semi-synthetic time series that can be used for that purpose. The tsMorph method gradually transforms a source time series into a target time series, generating a sequence of semi-synthetic time series. The data generated by tsMorph supports an empirical and systematic approach to understanding the behavior of algorithms.

The analysis of correlations between algorithm performance and meta-features revealed valuable insights into the strengths and weaknesses of the LSTM and DeepAR algorithms. Both algorithms demonstrated sensitivity to prediction accuracy, as indicated by strong negative correlations with the "Forecast Error" meta-feature. Additionally, their positive correlations with the "Centroid Frequency" meta-feature suggest proficiency in capturing long-term trends present in high-frequency components of the data. However, differences emerged between the algorithms regarding their responses to other meta-features, such as "Low Frequency Power" and "Whiten Timescale". 

The visual representation provided by the tsMorph performance understanding plot offers a comprehensive depiction of the relationships outlined in the correlation tables. This visualization highlights how variations in meta-feature values across morphing steps influence algorithm performance, providing a clearer understanding of the dynamics between meta-features and forecasting accuracy. By observing dispersion patterns and color gradients, researchers can gain deeper insights into the impact of meta-features on algorithm performance and identify trends that may not be immediately apparent from the correlation tables alone. Thus, the tsMorph performance understanding plot serves as a valuable tool for elucidating the intricate interplay between meta-features and algorithm performance, enhancing our comprehension of time series forecasting algorithms.

One limitation of the work presented here is that the transformation only applies to time series of the same size. However, we plan to develop other transformation options that can deal with time series of varying sizes with time series alignment techniques. Additionally, in the experiments, we focused on analyzing algorithms individually. However, by choosing the target and source time series differently, we can carry out other types of analyses (e.g., comparing the performance of two algorithms). In fact, tsMorph can be used to define and understand the borders of the meta-data space that delimit the areas of expertise of each algorithm. 

Finally, the semi-synthetic data generated by tsMorph can also be used as training data for AutoML and meta-learning approaches, addressing a major limitation of the current work in the area: the limited amount of meta-data.

\section*{Acknowledgments}
This work was partially funded by grants \#2019/10012-2 and \#2021/13281-4, São Paulo Research Foundation (FAPESP) and CNPq.
This work was partially funded also by AISym4Med (101095387) through the Horizon Europe Cluster 1: Health, ConnectedHealth (n.º 46858); Competitiveness and Internationalisation Operational Programme (POCI) and Lisbon Regional Operational Programme (LISBOA 2020), under the PORTUGAL 2020 Partnership Agreement, through the European Regional Development Fund (ERDF); NextGenAI - Center for Responsible AI (2022-C05i0102-02), supported by IAPMEI; FCT plurianual funding for 2020-2023 of LIACC (UIDB/00027/2020 UIDP/00027/2020). The computational resources of Google Cloud Platform were provided by the project CPCA-IAC/AF/594904/2023.

%
%
%
\bibliographystyle{splncs04}
\bibliography{references}
%




\end{document}